\documentclass[final]{cvpr}

\usepackage{times}
\usepackage{epsfig}
\usepackage{graphicx}
\usepackage{amsmath}
\usepackage{amssymb}
\usepackage{bm}

\usepackage[pagebackref=true,breaklinks=true,colorlinks,bookmarks=false]{hyperref}

\def\cvprPaperID{6889} 
\def\confYear{CVPR 2021}
\begin{document}

\makeatletter
\def\thanks#1{\protected@xdef\@thanks{\@thanks
        \protect\footnotetext{#1}}}
\makeatother

\title{ECKPN: Explicit Class Knowledge Propagation Network for Transductive Few-shot Learning}


\author{
		Chaofan Chen$^{1}$, \ Xiaoshan Yang$^{2,3}$,\  Changsheng Xu$^{2,3\dagger}$\thanks{$\dagger$ indicates corresponding author: Changsheng Xu. } , \ Xuhui Huang$^{4}$, \ Zhe Ma$^{4}$\\ 
		$^1$\begin{small}School of Information Science and Technology,\ \ University of Science and Technology of China (USTC)\end{small}  \\ 
			$^2$\begin{small}National Lab of Pattern Recognition (NLPR),\ \ Institute of Automation, Chinese Academy of Sciences (CASIA)\end{small}\\
			$^3$\begin{small}School of Artificial Intelligence,\ \ University of Chinese Academy of Sciences (UCAS)\end{small}\\
			$^4$\begin{small}X Lab, The Second Academy of CASIC, Beijing China\end{small}\\
	\tt\scriptsize chencfbupt@gmail.com, \tt\scriptsize \{xiaoshan.yang,csxu\}@nlpr.ia.ac.cn, \tt\scriptsize \{starhxh,mazhe\_thu\}@126.com
}
\maketitle

\pagestyle{empty}  
\thispagestyle{empty} 
\begin{abstract}
\vspace{-10pt}
Recently, 
the transductive graph-based methods
have achieved great success in the few-shot classification task. 
However, most existing methods ignore exploring the class-level knowledge that can be easily learned by humans from just a handful of samples.
In this paper, we propose an Explicit Class Knowledge Propagation Network (ECKPN), which is composed of the comparison, squeeze and calibration modules, 
to address this problem.
Specifically, we first employ the comparison module to explore the pairwise sample relations to learn rich sample representations
in the instance-level graph. 
Then, we squeeze the instance-level graph to generate the class-level graph, which can help obtain the class-level visual knowledge and facilitate modeling the relations of different classes. 
Next, the calibration module is adopted to characterize the relations of the classes explicitly to obtain the more discriminative class-level knowledge representations. Finally, we combine the class-level knowledge with the instance-level sample representations to guide the inference of the query samples. 
We conduct extensive experiments on four few-shot classification benchmarks, and the experimental results show that the proposed ECKPN significantly outperforms the state-of-the- art methods.
\end{abstract}
\vspace{-16pt}

\section{Introduction}\label{sec:intro}
\vspace{-6pt}
Recent deep learning methods rely on a large amount of labeled data to achieve high performance, 
which may have problems in some scenarios, 
where the cost of data collection is high, and thus it is difficult to obtain a large amount of labeled data.
The learning schema of these deep methods is different from that of humans.
%
After being exposed to a few data/samples, human beings can use their prior knowledge to learn quickly so as to successfully recognize new classes.
Therefore, how to reduce the gap between deep learning methods and human learning abilities has aroused the interest of many researchers.
Few-shot learning~\cite{fsl0,fsl1,fsl2}, which simulates the human learning schema,
has attracted much attention in the field of computer vision and machine learning.

\begin{figure}[t]
\centering
  \includegraphics[width=0.95\linewidth]{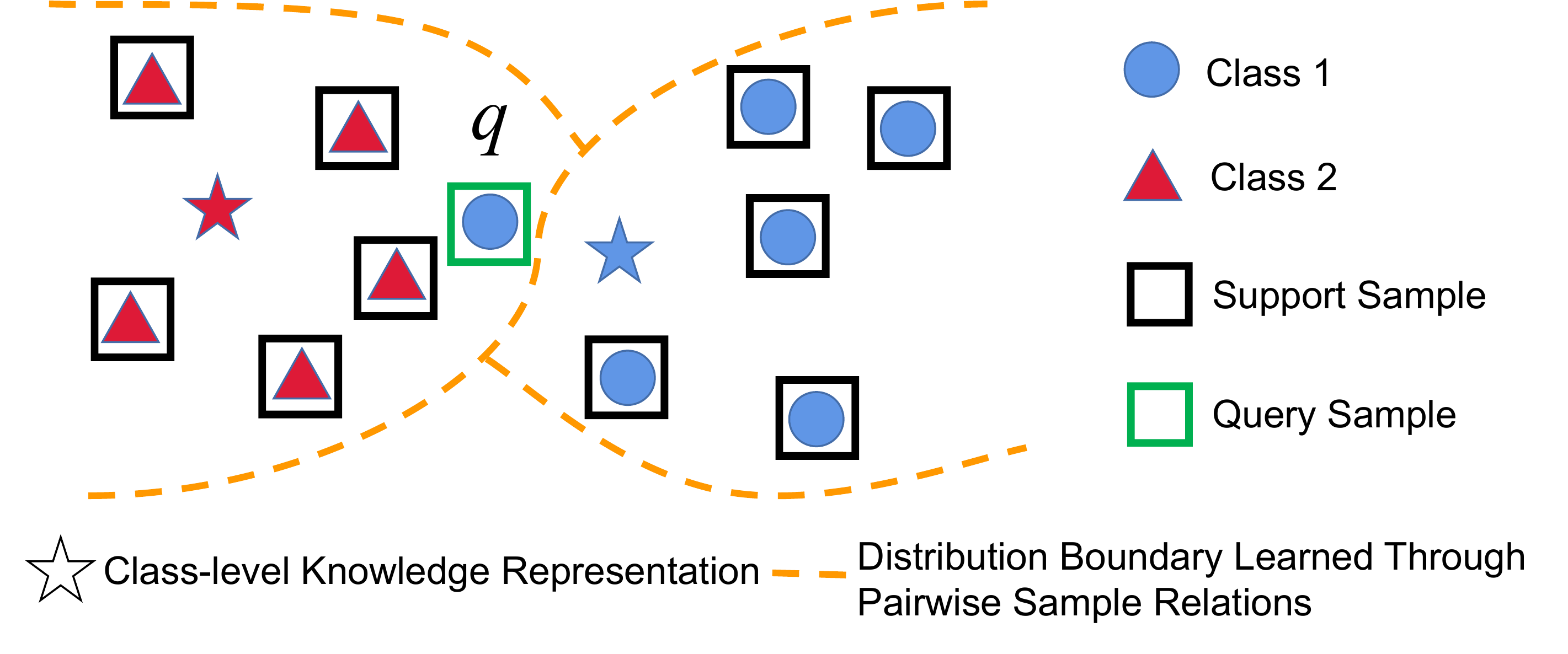}
  \caption{An illustration of the role of class-level knowledge representations (e.g., class centers).}
\label{fig:intro}
\end{figure}

\vspace{-2pt}
As a straightforward method to solve the few-shot learning task,
traditional fine-tuning techniques~\cite{ft}
can utilize the samples of the new classes to update the parameters of the network pretrained on the classes with sufficient samples.
However,
these methods always lead to over-fitting,
since only a few training samples
are not enough to represent the data distributions of the corresponding classes and learn effective classifiers.
%
%
A successful attempt to solve the over-fitting problem
is to apply the meta-learning mechanism~\cite{meta} in few-shot learning task.
The meta-learning based methods~\cite{maml,match,snail,reptile,afhn,leo,awgim,tpn,egnn,caml,deepemd,dpgn,metanas,trpn} are composed of two steps: meta-train and meta-test.
%
%
Each step (meta-train or meta-test) consists of multiple episodes (sub-tasks), and the data of each episode are composed of support set and query set.
%
%
These methods keep the meta-train environment consistent with the meta-test to help improve the generalization ability of the models,
thereby solving the problem of over-fitting.
Nowadays, meta-learning has become a general training mechanism in 
most of the few-shot learning methods.
%
In this paper, we also follow this training mechanism.
\vspace{-2pt}

Recently,
inspired by the success of graph networks in modeling structure information~\cite{gcn,graphsage,gat}, researchers began to propose the graph-based meta-learning approaches for few-shot learning and obtain the state-of-the-art performances~\cite{gnn-fsl,egnn,tpn,trpn,dpgn,graph-meta}.
These methods treat the samples as nodes to construct the graph and utilize the adjacency matrix to model the relations of images.
There are two settings of the graph-based meta-learning approaches: transductive setting and inductive setting.
The transductive methods characterize the relations of samples from both the support set and the query set for joint prediction, and thus obtain better performances than inductive methods, 
which can only learn a network based on the relations of support samples and classify each query sample individually.

Existing transductive graph-based methods learn to propagate the class label from the support set to the query set
by comprehensively considering the instance-level relations of samples.
%
However, these methods ignore the global context knowledge from
the perspective of a category.
%
In contrast,
people can learn richer representations of a new category from just a handful of samples, using them for creating new exemplars, and even creating new abstract categories based on existing categories~\cite{science}.
%
%
\textbf{
This inspires us to consider how to explicitly learn the richer class knowledge to guide the graph-based inference of query samples.
}
As illustrated in Figure~\ref{fig:intro}, if we only utilize the sample representations and relations to conduct the few-shot classification task, we may misclassify the query sample $q$ into class 2. However, if we learn the class-level knowledge representations explicitly to guide the inference procedure, we can classify $q$ correctly, because $q$ is closer to the representation of the class 1.

In order to address the above problem,
we propose an end-to-end transductive graph neural network,
which is called
Explicit Class Knowledge Propagation Network (ECKPN).
The proposed ECKPN is composed of the \textbf{comparison}, \textbf{squeeze} and \textbf{calibration} modules, which can be flexibly stacked \textbf{to
explicitly
learn and propagate the class-level knowledge.
}
%
%
%
%
%
\textbf{(1)} Firstly, the comparison module captures the rich representations of samples based on the pairwise relations in a instance-level graph.
The visual features are always structured vectors and many factors
(e.g., frequency, shapes,
illumination, textures)
could lead to grouping~\cite{ijcv,group} (i.e., a group of dimensions represents a semantic aspect or a piece of knowledge).
Thus,
we adopt multi-head relations in the message passing of the comparison module to characterize the group-wise relations of samples,
which provides fine-grained comparison of different samples.
Each node feature is divided into groups along the dimension,
and adjacency matrices are computed for different groups to obtain
multiple relation measurements, which are then aggregated to compute
the new node features of the samples.
%
\textbf{(2)} Then,
the squeeze module explores the intra-class context knowledge by
clustering samples with similar features from the instance-level graph, which results in a class-level graph.
%
The number of nodes in the class-level graph is same as the total number of classes.
Thus, each node represents the visual knowledge of a specific class.
\textbf{(3)}
Finally, the calibration module explicitly captures the relationships between different classes and learn more discriminative class knowledge to guide the graph-based inference of query samples.
%
Since the word embeddings of the class names
can provide rich semantic knowledge that
may not be contained in the visual contents,
we combine
them with the visual knowledge to obtain the multi-modal knowledge representations of different classes.
Based on the multi-modal knowledge representations,
a class-level message passing is adopted to exploit the relationship of different classes.
%
%
The new class-level knowledge representations obtained by message passing are combined with the corresponding instance-level sample representations
to guide the inference of the query samples.

To sum up, the main contributions of this paper are four-fold:

\begin{itemize}
\vspace{-6pt}
\item 
%
To the best of our knowledge, we are the first to propose an end-to-end graph-based few-shot learning architecture, which can explicitly learn the rich class knowledge to guide the graph-based inference of query samples.
\vspace{-6pt}
\item 
We build multi-head sample relations to explore the fine-grained comparison of pairwise samples, which can facilitate the learning of richer class knowledge based on the pairwise relations.
\vspace{-6pt}
\item
We leverage the semantic embeddings of the class names to construct the multi-modal knowledge representations of different classes,
which can provide more discriminative knowledge to guide the inference of the query samples.
\vspace{-6pt}
\item 
We conduct extensive experiments on four benchmarks (i.e., miniImageNet, tieredImageNet, CIFAR-FS and CUB-200-2011) for the transductive few-shot classification task, and the results show that the proposed method achieves the state-of-the-art performances. 
\end{itemize}

\vspace{-8pt}

\section{Related Work}
\vspace{-6pt}
In recent years, researchers have proposed many novel approaches to address the few-shot learning problems and achieved great success. 
As illustrated in ~\cite{awgim},
we can divide the existing few-shot learning methods into two categories: gradient-based ~\cite{maml,leo,reptile,caml,snail,metaopt,metasgd,metatransfer,metagan,other1,other2,other3} and metric-based~\cite{relation,proto,match,gnn-fsl,deepemd,tpn,trpn,afhn,dpgn,graph-meta,egnn,awgim,tapnet,tadam,covar}. 

{\bf Gradient-based Approaches.}
These approaches try to adapt to new classes within a few optimization steps.
The well-known model-agnostic meta-learning~\cite{maml} (MAML) method relies on the meta-learner~\cite{meta} to realize the fine-tune updates.
Reptile~\cite{reptile} is a first-order gradient-based meta-learning approach, which points out that MAML can be simply implemented.
It is trained on the sampled tasks and does not need a training-testing split for each task. 
Latent embedding optimization~\cite{leo} (LEO) is an encoder-decoder architecture, which utilizes the encoder to explore the low-dimensional latent embedding space for updating the representations and the decoder to predict the high-dimensional parameters.
Conditional class-aware meta-learning~\cite{caml} (CAML) conditionally transforms embeddings to explore the inter-class dependencies.
%
However, these gradient-based approaches usually fail to learn the effective sample representations for inference.
\vspace{-2pt}

{\bf Metric-based Approaches.} 
%
These methods usually embed the support and query samples into the same feature space at first, and then compute the similarity of features for prediction. 
Relation Networks~\cite{relation} exploit the pair-wise relations between support samples and query samples using the distance metric network.
Matching Networks~\cite{match} combine the attention mechanism and memory together to present an end-to-end differentiable nearest-neighbor classifier.
Prototypical Networks~\cite{proto} leverage the mean of the sample features of each class to build the prototype representations at first, and then compute the similarity between the query sample representation and the prototype representation for inference.
%
%
Recently, task-dependent adaptive metric~\cite{tadam} (TADAM) and task-adaptive projection network~\cite{tapnet} (TapNet) have been proposed to explore the task-dependent metric space to enhance the performance of existing few-shot models.
\vspace{-2pt}

The core of the metric-based approaches is exploring the relations between the query samples and the support samples/classes.
Inspired by the success of graph neural networks (GNNs)~\cite{gcn,graphsage,gat} on modeling the relationships and propagating information among points, researchers proposed many graph-based methods~\cite{gnn-fsl,egnn,tpn,trpn,dpgn,graph-meta} to conduct few-shot learning tasks and
have achieved great success. 
For example, GNN-FSL~\cite{gnn-fsl} is the first work to build an end-to-end trainable graph neural network architecture to conduct the few-shot classification task.
Transductive Propagation Network~\cite{tpn} (TPN) is the first to employ the GNNs to conduct the transductive inference.
It utilizes a closed-form solution to perform iterative label propagation.
Edge-Labeling Graph Neural Network~\cite{egnn} (EGNN) exploits the similarity/dissimilarity between nodes to dynamically update edge-labels.
Transductive relation-propagation graph neural network (TRPN) explicitly considers the relations of support-query pairs for few-shot learning.
%
Recent distribution propagation graph network (DPGN)~\cite{dpgn} builds a dual graph to model the distribution-level relations of samples and outperforms most existing methods in the classification task.
%
However, existing graph-based methods ignore exploring the class-level knowledge explicitly, which may limit their inference ability as illustrated in Figure ~\ref{fig:intro}.

\begin{figure*}[t]
\centering
   \includegraphics[width=0.85\linewidth]{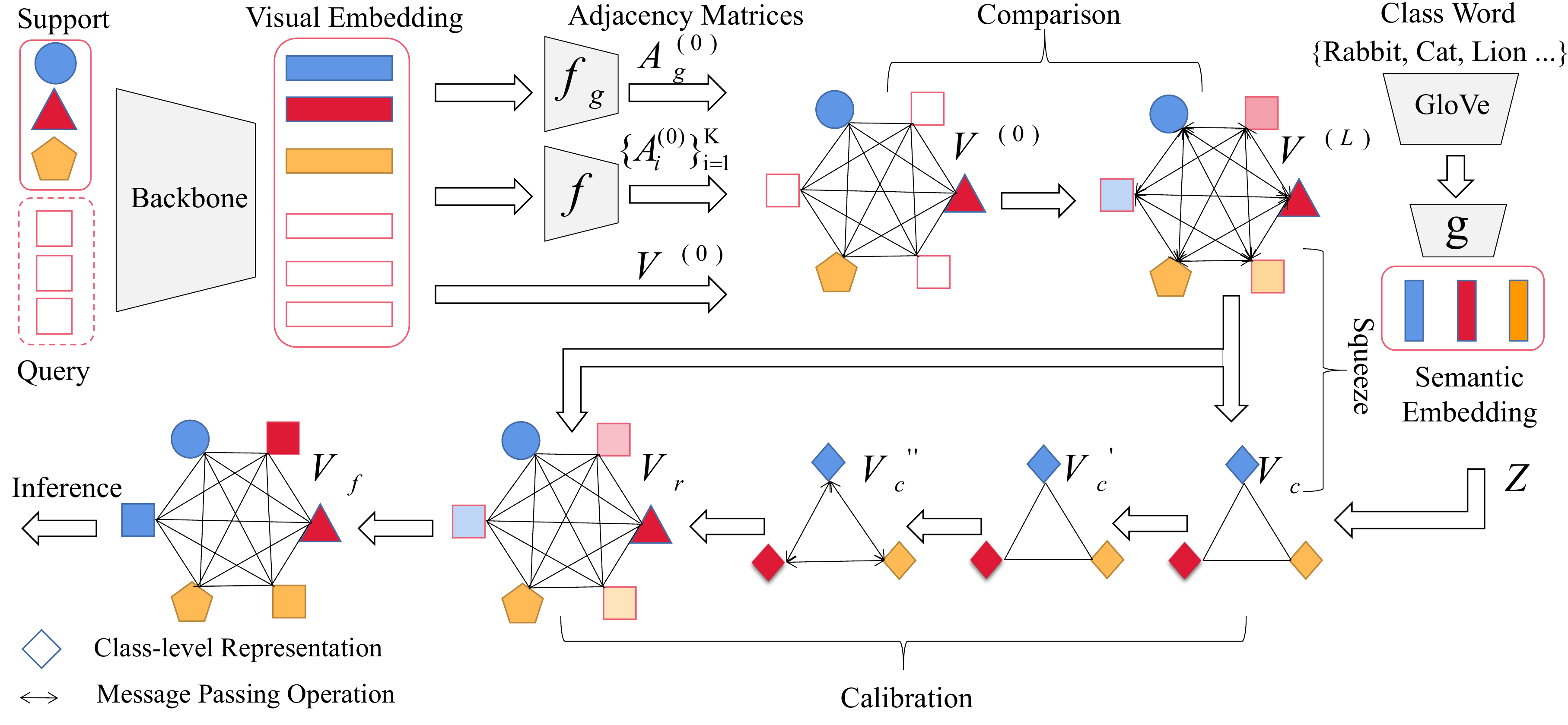}
   \caption{The overall framework of our proposed ECKPN. We take the 3-way 1-shot classification task as an example in this figure. Our ECKPN is composed of the comparison, squeeze and calibration modules, which can learn  and  propagate the class-level knowledge explicitly. Note that our comparison module contains $L$ message passing layers, but we just illustrate one layer for simplicity. }
   \vspace{-10pt}
\label{fig:architecture}
\end{figure*}

\vspace{-6pt}
\section{Method}\label{sec:method}
\vspace{-6pt}
\subsection{Problem Statement}
\vspace{-4pt}
As illustrated in Section~\ref{sec:intro}, we utilize the meta-learning mechanism to conduct the few-shot classification task.
For each episode in the meta-train,
we sample $N$ classes from $C_{train}$ (the class set of the training data $D_{train}$) to construct the support and query set.
The support set $S \subset  D_{train}$ contains $K$ samples for each class (i.e., the $N$-way $K$-shot setting), which can be denoted as $S=\{(x_1,y_1), (x_2,y_2), ... , (x_{N \times K},y_{N \times K})\}$, where $x_i$ represents the $i$-th sample and $y_i$ denotes the label of $x_i$.
The query set includes $T$ samples from the $N$ classes in total, which can be denoted as $Q =\{(x_{N \times K + 1}, y_{N \times K + 1}), ... , (x_{N \times K+T}, y_{N \times K + T})\}$.

For the transductive setting, 
we need to train a classification model which can leverage the $N \times K$ labeled support samples and the $T$ unlabeled query samples to
correctly predict the labels of the $T$ query samples.
%
The training procedure is employed episode by episode until convergence.
Given the test data set $D_{test}$ and its corresponding class set $C_{test}$, we construct the support and query set for the episode (in the meta-test) in a similar way as in the meta-train.
Note that $C_{train} \cap C_{test} = \varnothing $.
In the meta-test, we utilize the model learned in the meta-train to predict the labels of the query set samples.
The prediction/classification results are used to estimate the effectiveness of the model.

\textbf{Notations.} 
In this paper, $X_{i;m}$ denotes the $m$-th row of the matrix $X_i$
and $X_{i;m,n}$ denotes the element located in the $m$-th row and $n$-th column of the matrix $X_i$.

\subsection{Explicit Class Knowledge Propagation Network}\label{subsec:ECKPN}
\vspace{-6pt}
%
As illustrated in Figure~\ref{fig:architecture},
we first utilize the support and query samples to build an instance-level graph. 
Then, we leverage the comparison module to update the sample representations based on the pairwise node relations in the instance-level graph.
In this module, we construct the multi-head relations to help model the fine-grained relations of the samples to learn rich sample representations.
Next, we squeeze the instance-level graph to the class-level graph to explore the class-level visual knowledge explicitly.
In the calibration module, we perform the class-level message passing operation based on the relationships of the classes to update the class-level knowledge representations.
Since the semantic word embeddings of the classes can provide rich prior knowledge,
we combine them with the class-level visual knowledge to construct the multi-modal class knowledge representations
before the message passing of the calibration module.
Finally, the class-level knowledge representations are combined with the instance-level sample representations to guide the inference of the query samples.
\vspace{-10pt}
\subsubsection{Comparison Module: Instance-level Message Passing with Multi-head Relations}\label{subsubsec:comparison}
\vspace{-6pt}

For an image $i$, we employ a deep CNN model as backbone to extract its visual feature.
(We follow the existing works~\cite{gnn-fsl,dpgn} to combine the image visual features with its corresponding one-hot coding as the initial node features $v_i^{(0)} \in R^{d}$. Since the label of the query sample is not available in the inference produce, we set the elements of its one-hot coding to $1/N$, where $N$ is the number of classes.)
In each episode, we treat the support and query set samples as nodes to build the graph $G=(V^{(0)}, A^{(0)})$, where $V^{(0)}$ is the initial node feature matrix and $A^{(0)}$ is the initial adjacency matrix set which represents the sample relations. 
As illustrated in ~\cite{ijcv,group}, the visual features always contain some concepts that could lead to grouping, i.e., the feature dimensions from the same group represent similar knowledge.
However, existing graph-based few-shot learning methods usually directly utilize the global visual features to compute the similarities of the samples to construct the adjacency matrix, which cannot characterize the fine-grained relations well.
%
%
In this paper, we separate the visual features into $K$ chunks (i.e., $V^{(l)} = [V_{1}^{(l)}, V_{2}^{(l)}, ... ,V_{K}^{(l)}] \in R^{r \times d}$) and compute the similarities in each chunk to explore the multi-head relations of samples (i.e., $K$ adjacency matrices $A_1^{(l)}, A_2^{(l)}, ... , A_K^{(l)} \in R^{r \times r}$), where $r$ denotes the number of samples in each episode, $[*,*]$ denotes the concatenation operation and $l$ denotes that the matrix is generated in the $l$-th graph layer. Note that each chunk $V_i^{(l)}$ has the dimension of $d/K$. 
%
We also compute the global relation matrix $A_g^{(l)} \in R^{r \times r}$ based on the unchunked visual features.
We utilize the global ($A_g^{(l)}$)
and multi-head ($\{A_i^{(l)}\}_{i=1}^{K}$) relations 
jointly (i.e., $A^{(l)}=\{  A_g^{(l)}, A_1^{(l)}, ... ,A_K^{(l)}\}$) to propagate the information in the instance-level graph to update the sample representations.
In this way, we can explore the relations of samples more sufficiently and learn richer sample representations.
In the $l$-th layer, we leverage the updated sample representations $V^{(l)}$ to construct the new adjacency matrices $A_g^{(l)}$ and $A_i^{(l)}$ as follows:
\begin{equation}\label{eq:equ3}
\small
\setlength{\abovedisplayskip}{3pt}
\setlength{\belowdisplayskip}{3pt}
A_{g;m,n}^{(l)} = f_g((V_m^{(l)}-V_n^{(l)})^{2}), \quad A_{i;m,n}^{(l)} = f_i((V_{i;m}^{(l)}-V_{i;n}^{(l)})^{2})
\end{equation}
where $V_m$ denotes the visual feature of the $m$-th image, $V_{i;m}$ denotes the $i$-th chunk of $V_m$, and $(*)^{2}$ denotes the element-wise square operation. $f_i: R^{d/K} \rightarrow R^1$ and $f_g: R^{d} \rightarrow R^1$ are the mapping functions.

Inspired by the recent success of TRPN~\cite{trpn} in the few-shot classification task, we utilize the following matrix to mask the adjacency matrix:
\begin{equation}\label{eq:equ1}
\small
\setlength{\abovedisplayskip}{3pt}
\setlength{\belowdisplayskip}{3pt}
M_{m,n}=\left\{
\begin{array}{rcl}
-1 & & {if \ m,n \in S \ and \ y_m \ \neq \ y_n}\\
1 & & {otherwise}
\end{array} \right.
\end{equation}
where $m$ and $n$ are the samples in $S \cup Q$ and $y_m$ is the label of sample $m$.
%
%
This ensures that, for two samples from different categories, the higher the feature similarity of them, the more commonality decreases in the message passing process.
For the two samples from the same category, the results are exactly the opposite.

In the $l$-th layer, we utilize the $A^{(l-1)}$, $V^{(l-1)}$ and $M$ to generate the $V^{(l)}$ as follows:
\begin{equation}\label{eq:equ2}
\small
\setlength{\abovedisplayskip}{3pt}
\setlength{\belowdisplayskip}{3pt}
V^{(l)} = Tr([\parallel_{i=1}^{K}((A_i^{(l-1)} \odot  M)V_i^{(l-1)}), (A_g^{(l-1)} \odot  M)V^{(l-1)}])
\end{equation}
where 
$\parallel$ denotes the concatenation operation, 
$\odot $ denotes the element-wise multiplication operation and $Tr$ denotes the transformation function: $R^{r \times 2d} \rightarrow R^{r \times d}$. 
%
We repeat the above message passing $L$ times and obtain the new sample features $V^{(L)}$ which will be used in the squeeze module.
%
\vspace{-6pt}
\subsubsection{Squeeze Module: Class-level Visual Knowledge Learning}\label{subsubsec:squeeze}
\vspace{-6pt}
%

In order to obtain the class-level knowledge representations, we squeeze the instance-level graph to generate the class-level graph, where the nodes represent the visual knowledge of the classes.
For instance, we squeeze the nodes in the instance-level graph into 5 clusters/nodes so as to obtain the visual knowledge of the classes for the 5-way classification task. 
%
Specifically, we first utilize the ground truth to supervise the assignment matrix generation, and then squeeze samples according to the assignment matrix to obtain the class-level knowledge representations $V_c \in R^{r_1 \times d}$, where $r_1$ denotes the number of classes in each episode.

In this paper, we feed $V^{(L)}$ and $A_g^{(L)}$ into the standard graph neural network~\cite{gcn} to compute the assignment matrix $P \in R^{r \times r_1}$ for simplicity.
%
\begin{equation}\label{eq:equ5}
\small
\setlength{\abovedisplayskip}{3pt}
\setlength{\belowdisplayskip}{3pt}
P = softmax((A_g^{(L)} \odot M) V^{(L)} W)
\end{equation}
where $W \in R^{d \times r_1}$ denotes the trainable weight matrix and the $softmax$ operation is applied in a row-wise fashion.
Each element $P_{uv}$ in the assignment matrix $P$ represents the probability that node $u$ in the original graph is allocated to node $v$ in the class-level graph.
%
After generating the assignment matrix $P$, we utilize the following equation to generate the initial class-level knowledge representations:  
\begin{equation}\label{eq:equ6}
\small
\setlength{\abovedisplayskip}{3pt}
\setlength{\belowdisplayskip}{3pt}
V_c = P^TV^{(L)}
\end{equation}
where $T$ denotes transpose operation.
In the class-level graph, 
each node feature can be considered as the weighted sum of the node features with the same label in the instance-level graph.
%
%
In this way, we obtain the class-level visual knowledge representations, which will facilitate modeling the relations of different classes in the calibration module.
%
\vspace{-10pt}

\subsubsection{Calibration Module: Class-level Message Passing with Multi-modal Knowledge}
\vspace{-6pt}
%
Since the class word embeddings can provide the information that may not be contained in the visual content,
we combine them with the generated class-level visual knowledge to construct the multi-modal knowledge representations.
%
%
Specifically, we first leverage the GloVe (pretrained on a large text corpora with self-supervised constraint)~\cite{glove} to obtain the $d_1$-dimensional semantic embeddings of class labels.
The Common Crawl version of the GloVe is used in this paper, which is trained on 840B tokens.
More details can be found in~\cite{glove}.
After obtaining the word embeddding $e_i \in R^{d_1}$ of the $i$-th class, we employ a mapping network $g: R^{d_1} \rightarrow R^{d}$ to map it into a semantic space 
which has the same dimension with the visual knowledge representation, i.e., $z_i = g(e_i) \in R^{d}$.
Finally, we obtain the multi-modal class representations as follows:
\begin{equation}\label{eq:equ18}
\small
\setlength{\abovedisplayskip}{3pt}
\setlength{\belowdisplayskip}{3pt}
V_c^{'} = [V_c, Z] 
\end{equation}
where $Z \in R^{r_1 \times d}$ is the matrix of semantic word embeddings. 
%
%
In this way, we can obtain richer class-level knowledge representations.

The adjacency matrix ($A_c$) of the class-level graph represents the relations of the class representations and its value denotes the connectivity strength of the class pairs.
In this paper, we leverage the following equations to compute the adjacency matrix $A_c$ 
and the new class-level knowledge representations $V_c^{''}$.
%
\begin{equation}\label{eq:equ7}
\small
\setlength{\abovedisplayskip}{3pt}
\setlength{\belowdisplayskip}{3pt}
A_c = P^T A_g P, \quad V_c^{''} = A_c V_c^{'} W^{'} 
\end{equation}
where $W^{'} \in R^{2d \times 2d}$ is a trainable weight matrix.
%
In order to make each sample contain the corresponding class knowledge learned in~\eqref{eq:equ7},
we utilize the assignment matrix to map the class knowledge back to the instance-level graph as follows:
\begin{equation}\label{eq:equ9}
\small
\setlength{\abovedisplayskip}{3pt}
\setlength{\belowdisplayskip}{3pt}
V_r = PV_c^{''}
\end{equation}
where $V_r \in R^{r \times 2d}$ denotes the refined features.
Finally, we combine $V_r$ with $V^{(L)}$ by concatenation to generate the sample representations $V_f$ for query inference.

\subsection{Inference}\label{subsubsec:infer}
\vspace{-6pt}
To infer the class labels of query samples,
we utilize $V_f$ to compute the corresponding adjacency matrix $A_f$ as follows:
\begin{equation}\label{eq:equ17}
\small
\setlength{\abovedisplayskip}{3pt}
\setlength{\belowdisplayskip}{3pt}
A_{f;m,n} = f_l((V_{f;m}-V_{f;n})^{2})
\end{equation}
%
where $V_{f;m}$ and $V_{f;n}$ are the representations of the $m$-th sample and the $n$-th sample, respectively.
$f_l: R^{3d} \rightarrow R^1$ is a mapping function.
For each query sample, we leverage the class labels of the support samples to predict its label:
\begin{equation}\label{eq:equ13}
\small
\setlength{\abovedisplayskip}{3pt}
\setlength{\belowdisplayskip}{3pt}
\widetilde{y_v} = softmax\big(\sum_{u=1}^{N\times K}A_{f;u,v} \cdot one\text{-}hot(y_u)\big)
\end{equation}
where $one$-$hot$ denotes the one-hot encoder.
%

\subsection{Loss Function}
\vspace{-6pt}
The overall framework of the proposed ECKPN can be optimized in an end-to-end form by the following loss function:
\begin{equation}\label{eq:equ16}
\small
\setlength{\abovedisplayskip}{3pt}
\setlength{\belowdisplayskip}{3pt}
\mathcal{L} = \lambda_0 \mathcal{L}_0 + \lambda_1 \mathcal{L}_1 + \lambda_2 \mathcal{L}_2
\end{equation}
where $\lambda_0$, $\lambda_1$ and $\lambda_2$ are hyper-parameters that are set to 1.0, 0.5 and 1.0 in the experiment.
$\mathcal{L}_0$, $\mathcal{L}_1$ and $\mathcal{L}_2$ are adjacency loss, assignment  loss and classification loss respectively, that will be introduced as follows.

\textbf{Adjacency Loss:} As illustrated in Section~\ref{subsubsec:comparison},
for each graph network layer $l \in \{1, ..., L\}$ in the comparison module, we have multiple adjacency matrices $A_g^{(l)}$ and $\{A_i^{(l)} \}_{i=1}^{K}$ for message passing between support and
 query samples.
In addition, we have the adjacency matrix $A_f$ for query inference  in Section~\ref{subsubsec:infer}.
To ensure these adjacency matrices to be able to capture the correct sample relations, we use the following loss function:
%
\begin{equation}\label{eq:equ10}
\footnotesize
\setlength{\abovedisplayskip}{3pt}
\setlength{\belowdisplayskip}{3pt}
\mathcal{L}_0 = - \sum_{A_* \in A_s} ( \frac{sum(log(A_*) H G_t)}{sum(H G_t)} + \frac{sum(log(1-A_*) H (1-G_t))}{sum(H (1-G_t))})
\end{equation}
%
where $A_s = \{A_g^{(1)}, ... , A_g^{(L)}\} \cup \{A_f\}\cup \{A_i^{(1)}, ... , A_i^{(L)} \}_{i=1}^{K}$ and $sum(*)$ denotes the sum of all elements in the matrix.
$H \in R^{r \times r}$ is the query mask and $G_t \in R^{r \times r}$ is the ground truth matrix which are defined as follows:
%
\begin{equation}\label{eq:equ11}
\small
\setlength{\abovedisplayskip}{3pt}
\setlength{\belowdisplayskip}{3pt}
H_{m,n}=\left\{
\begin{array}{rl}
0 & {if \ m \in S}\\
1 & {otherwise}
\end{array} \right.,\quad
G_{t;m,n}=\left\{
\begin{array}{rl}
1 & {if \ y_m = y_n}\\
0 & {otherwise}
\end{array} \right.
\end{equation}
where $m$ and $n$ denote the nodes in the graph.
%
%
%

%
\textbf{Assignment Loss:} To ensure that the assignment matrix $P$ computed in the squeeze module (illustrated in Section~\ref{subsubsec:squeeze}) can correctly cluster the samples with the same label, we utilize the following cross-entropy loss function:
%
\begin{equation}\label{eq:equ15}
\small
\setlength{\abovedisplayskip}{3pt}
\setlength{\belowdisplayskip}{3pt}
\mathcal{L}_1 = \mathcal{L}_{ce}(P, one\text{-}hot\big([C_s, C_q])\big)
\end{equation}
%
where $one\text{-}hot([C_s, C_q])$ denotes the ground truth one-hot class vectors of the support and query samples.

%
\textbf{Classification Loss:} To constrain that the proposed ECKPN can predict the correct query labels, we use the following loss function:
\begin{equation}\label{eq:equ14}
\small
\setlength{\abovedisplayskip}{3pt}
\setlength{\belowdisplayskip}{3pt}
\mathcal{L}_2 = \sum_{v \in Q}\mathcal{L}_{ce}(\widetilde{y_v}, y_v)
\end{equation}
where $\mathcal{L}_{ce}$ denotes the cross-entropy loss function.

\section{Experiments}
%

\subsection{Datasets}
\vspace{-6pt}
MiniImageNet~\cite{match} and tieredImageNet~\cite{proto1} are two popular few-shot benchmarks derived from the ILSVRC-12 dataset~\cite{ijcv1}.
The miniImageNet contains 100 classes with 600 images per class.
Each image is RGB-colored and has the size of $84 \times 84$.
The tieredImageNet contains 779165 images of size $84 \times 84$ sampled from 608 classes.
CIFAR-FS~\cite{cifar} is reorganized from the CIFAR-100 dataset for the few-shot classification task.
It contains 100 classes with 60000 images in total.
Each image has the size of $32 \times 32$.
CUB-200-2011~\cite{cub} is a medium-scale dataset used for fine-grained classification.
It has 11788 images of size $84 \times 84$ from 200 bird categories.
We follow the popularly used train/val/test setting proposed in~\cite{proto1,ijcv1,cifar,dpgn,trpn}.
The statistics of these benchmarks are shown in Table ~\ref{tab1}.
\vspace{-6pt}

\subsection{Experimental Setup}
\vspace{-6pt}
{\bf Architectures.} We utilize two popular backbones (Conv-4~\cite{maml,egnn} and ResNet-12~\cite{resnet,snail,dpgn}) to encode the input images into 128 dimensions.
Both Conv-4 and ResNet-12 consist of four blocks.
Each block in Conv-4 is composed of $3 \times 3$ convolutions, a batch normalization (BN) and a  LeakyReLU activation.
Each residual block in ResNet-12 contains 3 convolutional layers with the size of $3 \times 3$.
Each convolutional layer is followed by a $2 \times 2$ max-pooling layer.
A global average-pooling is applied in the end of the fourth block.
Before feeding the images to the backbones, we follow the recent few-shot learning approaches~\cite{data0,data1,dpgn} to perform data augmentation, i.e., color jittering, random crop and horizontal flip. Note that the mapping and transformation functions $f_i$, $f_g$, $f_l$ and $T_r$ are single-layer convolutional networks with batch normalization and LeakyReLU.

{\bf Training.}
We train our model on miniImageNet, tieredImageNet, CIFAR-FS and CUB-200-2011 for 200K, 200K, 100K and 100K iterations respectively.
In each iteration, we construct 28 episodes for meta-training.
Adam optimizer ~\cite{adam} is used in all experiments with the initial learning rate 0.001.
%
We set the weight decay to 1e-5 and decay the learning rate by 0.1 every 15K iterations.
\vspace{0pt}

{\bf Evaluation.}
We conduct the 5-way 1-shot and 5-shot experiments on the four benchmarks for few-shot classification task.
%
We follow~\cite{dpgn,trpn} to construct 10K episodes in the meta-test and report the mean prediction accuracy of them to measure the effectiveness of the proposed ECKPN.

\begin{table}
\small
\centering
\begin{tabular}{lcccc}
\hline
Dataset & Classes & Images &Train/Val/Test\\
\hline\hline
miniImageNet & 100 &60000 &64/16/20\\\hline
tieredImageNet & 608 &779165 &351/97/160\\\hline
CIFAR-FS & 100 & 60000 &64/16/20\\\hline
CUB-200-2011 &200 & 11788 &100/50/50\\\hline
\end{tabular}
\caption{The statistics of the four few-shot benchmarks.}
\label{tab1}
\end{table}

\subsection{Classification Results}
\vspace{-6pt}
We compare the classification results of the proposed ECKPN with the recent state-of-the-art few-shot methods and report the classification results of the 5-way 1-shot and 5-shot under different backbones (Conv-4 and ResNet-12) in Table~\ref{tab2},~\ref{tab3} and~\ref{tab5}.
%
%
From these experimental results, we have the following observations.
(1) The proposed ECKPN achieves the state-of-the-art classification results compared with the recent methods on all four benchmarks for both the 5-shot and 1-shot setting, which demonstrates the effectiveness of our model.
Especially for the 1-shot setting on the miniImageNet dataset,
the proposed method equipped with the Conv-4 and ResNet-12 achieves improvement of 2.88\% and 2.71\% respectively compared with the second-best approach DPGN.
These results demonstrate the necessity of modeling the class-level knowledge in the few-shot classification task.
(2) The proposed method achieves more improvements in the 1-shot setting than in the 5-shot setting.
Since the number of samples in the 5-shot setting is larger than in the 1-shot setting, with more samples, the recent graph-based methods can adapt to the novel classes better based on only the sample relations, which reduces the performance gain of our ECKPN.
However, our ECKPN can still achieve the improvements of 0.7\%-0.8\% under the 5-shot setting on all four benchmarks.
\vspace{-6pt}

\begin{table}
\small
\centering
\setlength{\tabcolsep}{4pt}
\begin{tabular}{lccc}
\hline
Method  & Backbone & 5way-1shot &5way-5shot\\
\hline
MatchingNet~\cite{match} & Conv-4 &$43.56$ \scriptsize{$\pm0.84$} &$55.31$ \scriptsize{$\pm0.73$}  \\
ProtoNet~\cite{proto} &Conv-4 &$49.42$ \scriptsize{$\pm0.78$} &$68.20$ \scriptsize{$\pm0.66$} \\
RelationNet~\cite{relation} &Conv-4 &$50.44$ \scriptsize{$\pm 0.82$} &$65.32$ \scriptsize{$\pm0.70$} \\
Dynamic~\cite{data0} &Conv-4 &$56.20$ \scriptsize{$\pm0.86$} &$71.94$ \scriptsize{$\pm0.57$} \\
Reptile~\cite{reptile} &Conv-4 &$49.97$ \scriptsize{$\pm0.32$} &$65.99$ \scriptsize{$\pm0.58$} \\
MAML~\cite{maml} &Conv-4 &$48.70$ \scriptsize{$\pm 1.84$} &$55.31$ \scriptsize{$\pm0.73$} \\
Meta-SGD~\cite{metasgd} &Conv-4 &$50.47$ \scriptsize{$\pm 1.87$} &$64.03$ \scriptsize{$\pm0.94$ }\\
GNN-FSL~\cite{gnn-fsl} &Conv-4 &$50.33$ \scriptsize{$\pm 0.36$} &$66.41 $ \scriptsize{$\pm0.63$} \\
TPN~\cite{tpn} &Conv-4 &$55.51$ \scriptsize{$\pm0.86$} &$69.86$ \scriptsize{$\pm 0.65$} \\
EGNN~\cite{egnn} &Conv-4 &- &$76.34$ \scriptsize{$\pm0.48$} \\
TRPN~\cite{trpn} &Conv-4 &$57.84$ \scriptsize{$\pm 0.51$} &$78.57$ \scriptsize{$\pm0.44$}\\
DPGN~\cite{dpgn} &Conv-4 &$66.01$ \scriptsize{$\pm 0.36$ }&$82.83$ \scriptsize{$\pm0.41$}\\
\textbf{ECKPN} &\textbf{Conv-4} &\bm{$68.89 $ }\scriptsize{$\pm 0.34$} &\bm{$83.59 $ }\scriptsize{$\pm 0.44$} \\\hline
LEO~\cite{leo} &Others &$61.76 $ \scriptsize{$\pm 0.08$}  &$77.59 $ \scriptsize{$\pm 0.12$}\\
CloserLook~\cite{close} &Others &$51.75 $ \scriptsize{$\pm 0.80$}  &$74.27 $ \scriptsize{$\pm 0.63$}\\
CTM~\cite{ctm} &Others &$62.05 $ \scriptsize{$\pm 0.55$ } &$78.63 $ \scriptsize{$\pm 0.06$}\\
wDAE~\cite{wdae} &Others &$61.07 $ \scriptsize{$\pm 0.15$}  &$76.75 $ \scriptsize{$\pm 0.11$}\\
AWGIM~\cite{awgim} &Ohers &$63.12 $ \scriptsize{$\pm 0.08$} &$78.40 $ \scriptsize{$\pm 0.11$}\\
AFHN ~\cite{afhn} &Others &$62.38 $ \scriptsize{$\pm 0.72$} &$78.16 $ \scriptsize{$\pm 0.56$}\\
\hline
FEAT~\cite{data1} &ResNet-12 &$62.96 $ \scriptsize{$\pm 0.02$}  &$78.49 $ \scriptsize{$\pm 0.02$}\\
TADAM~\cite{tadam} &ResNet-12 &$58.50$ \scriptsize{$\pm0.30$} &$76.70$ \scriptsize{$\pm 0.30$} \\
TapNet~\cite{tapnet} &ResNet-12 &$61.65$ \scriptsize{$\pm0.15$} &$76.36 $ \scriptsize{$\pm 0.10$} \\
MataGAN~\cite{metagan} &ResNet-12 &$52.71$ \scriptsize{$\pm 0.64$} &$68.63$ \scriptsize{$\pm 0.67$}\\
Shot-Free~\cite{short} &ResNet-12 &$59.04$ \scriptsize{$\pm 0.43$} &$77.64$ \scriptsize{$\pm 0.39$}\\
SNAIL~\cite{snail} &ResNet-12 &$55.71$ \scriptsize{$\pm0.99$} &$68.88$ \scriptsize{$\pm 0.92$} \\
MTL~\cite{metatransfer} &ResNet-12 &$61.20$ \scriptsize{$\pm1.80$} &$75.53$ \scriptsize{$\pm0.80$}\\
MetaOptNet~\cite{metaopt} &ResNet-12 &$62.64$ \scriptsize{$\pm0.61$} &$78.63 $ \scriptsize{$\pm 0.46$}\\
DeepEMD~\cite{deepemd} &ResNet-12 &$65.91$ \scriptsize{$ \pm 0.82$} &$82.41 $ \scriptsize{$\pm 0.56$} \\
DPGN~\cite{dpgn} &ResNet-12 &$67.77$ \scriptsize{$\pm0.32$} &$84.60$ \scriptsize{$\pm0.43$ }\\
\textbf{ECKPN} &\textbf{ResNet-12} &\bm{$70.48$ }\scriptsize{$\pm0.38$} &\bm{$85.42 $ }\scriptsize{$\pm 0.46$} \\\hline
\end{tabular}
\caption{Few-shot classification accuracies (\%) on miniImageNet.}
\label{tab2}
\end{table}

\subsection{Semi-supervised Classification Results}
\begin{figure}[t]
\centering
   \includegraphics[width=1.0\linewidth]{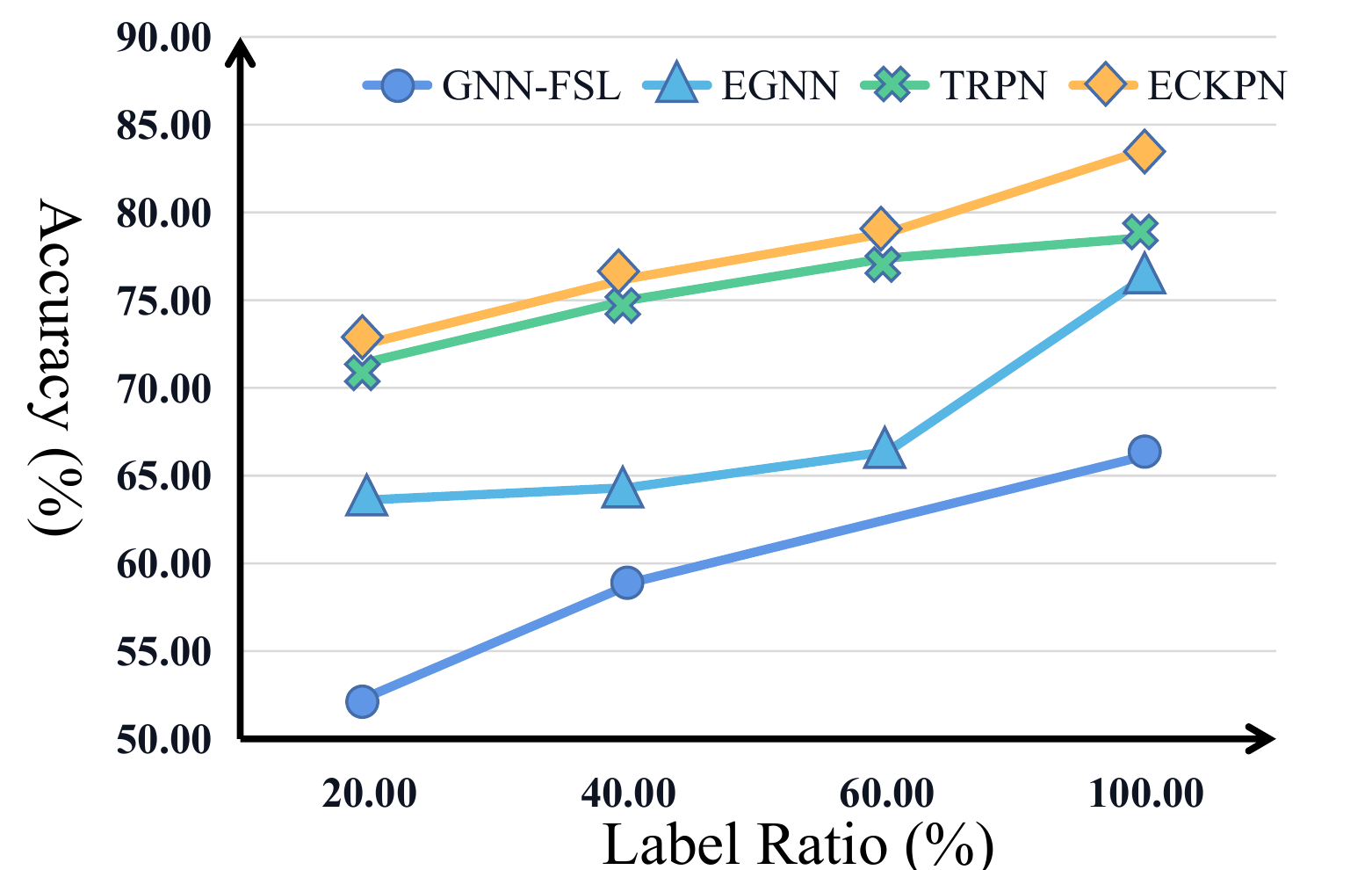}
   \caption{Semi-supervised few-shot classification accuracies (\%) in 5-way 5-shot on miniImageNet.}
\label{fig:exp1}
\end{figure}

\vspace{-6pt}
In this part, we apply the proposed ECKPN in the semi-supervised classification task to further evaluate its generalization ability.
Specifically, we follow ~\cite{egnn,tpn} to partially label the support samples with different ratios (i.e., 20\%, 40\%, 60\% and 100\%).
The label ratio 20\% means that 20\% labeled and 80\% unlabeled support samples are used to train the model in each episode. 
We compare the proposed ECKPN with the recent GNN-FSL~\cite{gnn-fsl}, EGNN~\cite{egnn} and TRPN~\cite{trpn} equipped with Conv-4.
We show the 5-way 5-shot classification results in Figure ~\ref{fig:exp1}.
As shown, the proposed ECPKN achieves better performances than existing methods under all label ratios, which demonstrates the effectiveness of capturing the class-level knowledge to guide the inference of the query samples.

\begin{table}
\small
\setlength{\tabcolsep}{4pt}
\begin{tabular}{lccc}
\hline
Method  & Backbone & 5way-1shot &5way-5shot\\
\hline
MatchingNet~\cite{match} & Conv-4  &$54.02$ \scriptsize{$\pm0.00$} &$70.11$ \scriptsize{$\pm 0.00$} \\
ProtoNet~\cite{proto} &Conv-4  &$53.31$ \scriptsize{$\pm0.89$} &$72.69$ \scriptsize{$\pm0.74$}\\
RelationNet~\cite{relation} &Conv-4  &$54.48$ \scriptsize{$\pm0.93$} &$71.32$ \scriptsize{$\pm0.70$}\\
Reptile~\cite{reptile} &Conv-4  &$52.36$ \scriptsize{$\pm0.23$} &$71.03$ \scriptsize{$\pm 0.22$}\\
MAML~\cite{maml} &Conv-4 &$51.67$ \scriptsize{$\pm1.81$} &$70.30$ \scriptsize{$\pm0.08$}\\
Meta-SGD~\cite{metasgd} &Conv-4  &$62.95$ \scriptsize{$\pm0.03$} &$79.34$ \scriptsize{$\pm0.06$}\\
GNN-FSL~\cite{gnn-fsl} &Conv-4  &$43.56$ \scriptsize{$\pm0.84$} &$55.31$ \scriptsize{$\pm0.73$}\\
TPN~\cite{tpn} &Conv-4  &$57.53$ \scriptsize{$\pm0.96$} &$72.85$ \scriptsize{$\pm0.74$}\\
EGNN~\cite{egnn} &Conv-4  &-&$80.15$ \scriptsize{$\pm0.30$}\\
TRPN~\cite{trpn} &Conv-4 &$59.26$ \scriptsize{$\pm0.50$} &$79.66$ \scriptsize{$\pm0.45$}\\
DPGN~\cite{dpgn} &Conv-4  &$69.43 $ \scriptsize{$\pm 0.49$} &$85.92$ \scriptsize{$\pm0.42$}\\
\textbf{ECKPN} &\textbf{Conv-4}  &\bm{$70.45 $} \scriptsize{$\pm 0.48$} &\bm{$86.74$ }\scriptsize{$ \pm 0.42$}\\
\hline
wDAE~\cite{wdae} &Others  &$68.18 $ \scriptsize{$\pm 0.16$} &$83.09$ \scriptsize{$\pm0.12$}\\
CTM~\cite{ctm} &Others  &$64.78 $ \scriptsize{$\pm 0.11$} &$81.05$ \scriptsize{$\pm0.13$}\\
LEO~\cite{leo} &Others  &$66.33 $ \scriptsize{$\pm 0.05$} &$81.44$ \scriptsize{$\pm0.09$}\\
AWGIM~\cite{awgim} &Others &$67.69 $ \scriptsize{$\pm 0.11$} &$82.82 $ \scriptsize{$\pm 0.13$}\\
\hline
MetaOptNet~\cite{metaopt} &ResNet-12 &$65.81 $ \scriptsize{$\pm0.74$} &$81.75 $ \scriptsize{$\pm0.53$}\\
TapNet~\cite{tapnet} &ResNet-12  &$63.08 $ \scriptsize{$\pm0.15$} &$80.26 $ \scriptsize{$\pm0.12$}\\
DeepEMD~\cite{deepemd} &ResNet-12 &$71.16$ \scriptsize{$\pm 0.87$} &$86.03$ \scriptsize{$\pm 0.58$}\\ 
Shot-Free~\cite{short} &ResNet-12 &$66.87 $ \scriptsize{$\pm0.43$} &$82.64 $ \scriptsize{$\pm0.39$}\\
DPGN~\cite{dpgn} &ResNet-12  &$72.45$ \scriptsize{$\pm0.51$} &$87.24$ \scriptsize{$\pm0.39$}\\
\textbf{ECKPN} &\textbf{ResNet-12}  &\bm{$73.59 $} \scriptsize{$\pm 0.45$} &\bm{$88.13 $} \scriptsize{$\pm 0.28$}\\\hline
\end{tabular}
\caption{Few-shot classification accuracies (\%) on tieredImageNet.}
\label{tab3}
\end{table}

\begin{figure}[t]
\centering
   \includegraphics[width=1.0\linewidth]{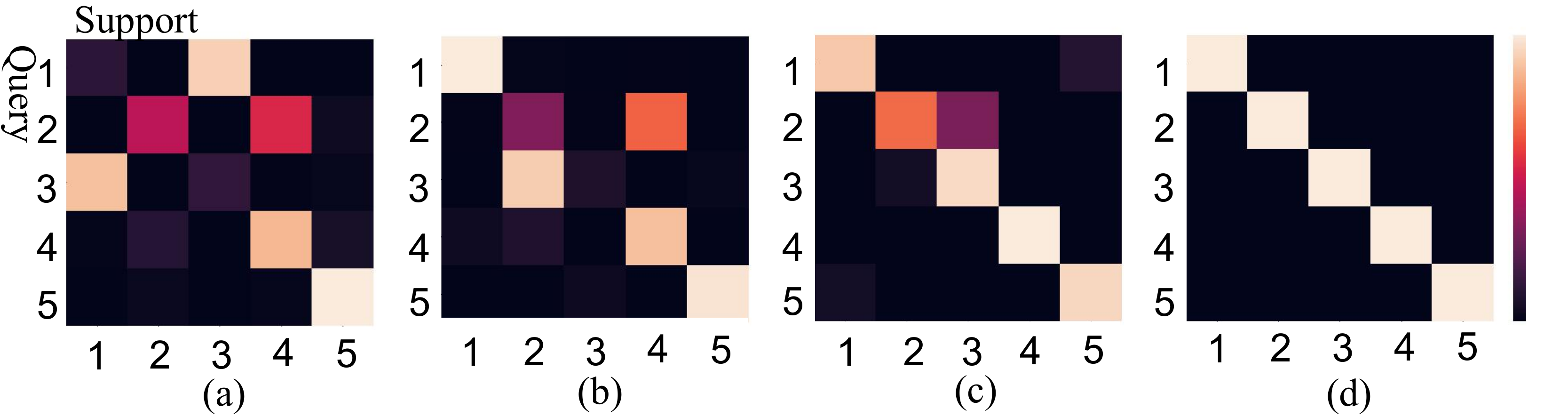}
   \caption{The visualization of the support-query similarities in 5-way 1-shot setting. (a), (b) and (c) represent the similarities of support and query samples in the first, third and last layers of the comparison module. (d) denotes the ground truth support-query similarities. The white denotes the high confidence and the black denotes the low confidence.}
\label{fig:exp0}
\end{figure}

\begin{table}
\small
\centering
\setlength{\tabcolsep}{4pt}
\begin{tabular}{lccc}
\hline
\multicolumn{4}{c}{CUB-200-2011}\\\hline
Method  & Backbone & 5way-1shot &5way-5shot\\
\hline
ProtoNet~\cite{proto} &Conv-4  &$51.31$ \scriptsize{$\pm0.91$} &$70.77$ \scriptsize{$\pm0.69$}\\
RelationNet~\cite{relation} &Conv-4  &$62.45$ \scriptsize{$\pm0.98$} &$76.11$ \scriptsize{$\pm0.69$}\\
MatchingNet~\cite{match} &Conv-4 &$61.16$ \scriptsize{$\pm0.89$} &$72.86$ \scriptsize{$\pm0.70$}\\
MAML~\cite{maml} &Conv-4 &$55.92$ \scriptsize{$\pm0.95$} &$72.09$ \scriptsize{$\pm0.76$}\\
DN4~\cite{dn4} &Conv-4 &$53.15$ \scriptsize{$\pm0.84$} &$81.90$ \scriptsize{$\pm0.60$}\\
CloserLook~\cite{close} &Conv-4 &$60.53$ \scriptsize{$\pm0.83$} &$79.34$ \scriptsize{$\pm0.61$}\\
DPGN~\cite{dpgn} &Conv-4  &$76.05$ \scriptsize{$\pm0.51$} &$89.08$ \scriptsize{$\pm0.38$}\\
\textbf{ECKPN} &\textbf{Conv-4}  &\bm{$77.20 $} \scriptsize{$\pm 0.36$} &\bm{$89.72 $} \scriptsize{$\pm 0.31$}\\\hline
DeepEMD~\cite{deepemd} &ResNet-12 &$75.65$ \scriptsize{$\pm0.83$} &$88.69 $ \scriptsize{$\pm 0.50$}\\
TADAM &ResNet-12 &$72.00$ \scriptsize{$\pm0.70$} &$84.20$ \scriptsize{$\pm0.50$}\\
FEAT~\cite{data1} &ResNet-12 &$68.87$ \scriptsize{$\pm0.22$} &$82.90$ \scriptsize{$\pm0.15$}\\
DPGN~\cite{dpgn} &ResNet-12  &$75.71$ \scriptsize{$\pm0.47$} &$91.48$ \scriptsize{$\pm0.33$}\\
\textbf{ECKPN} &\textbf{ResNet-12}  &\bm{$77.43 $ }\scriptsize{$\pm 0.54$} &\bm{$92.21$} \scriptsize{$ \pm 0.41$}\\\hline
\multicolumn{4}{c}{CIFAR-FS}\\\hline
Method  & Backbone & 5way-1shot &5way-5shot\\
\hline
ProtoNet~\cite{proto} &Conv-4  &$55.5$ \scriptsize{$\pm0.7$} &$72.0$ \scriptsize{$\pm0.6$}\\
RelationNet~\cite{relation} &Conv-4  &$55.0$ \scriptsize{$\pm1.0$} &$69.3$ \scriptsize{$\pm0.8$}\\
MAML~\cite{maml} &Conv-4 &$58.9$ \scriptsize{$\pm1.9$} &$71.5$ \scriptsize{$\pm1.0$}\\
R2D2~\cite{cifar} &Conv-4 &$65.3$ \scriptsize{$\pm0.2$} &$79.4$ \scriptsize{$\pm0.1$}\\
DPGN~\cite{dpgn} &Conv-4  &$76.4$ \scriptsize{$\pm0.5$} &$88.4$ \scriptsize{$\pm0.4$}\\
\textbf{ECKPN} &\textbf{Conv-4} &\bm{$77.5$ }\scriptsize{$ \pm 0.4$} &\bm{$89.1$} \scriptsize{$\pm 0.5$}\\
\hline
DeepEMD~\cite{deepemd} &ResNet-12 &$46.47$ \scriptsize{$ \pm 0.8$} &$63.22 $ \scriptsize{$\pm 0.7$}\\
MetaOpNet~\cite{metaopt} &ResNet-12 &$72.0$ \scriptsize{$\pm0.7$} &$84.2$ \scriptsize{$\pm0.5$}\\
Shot-Free~\cite{short} &ResNet-12 &$69.2$ \scriptsize{$\pm0.4$} &$84.7$ \scriptsize{$\pm0.4$}\\
DPGN~\cite{dpgn} &ResNet-12  &$77.9$ \scriptsize{$\pm0.5$} &$90.2$ \scriptsize{$\pm0.4$}\\
\textbf{ECKPN} &\textbf{ResNet-12}  &\bm{$79.2 $ }\scriptsize{$\pm 0.4$} &\bm{$91.0$} \scriptsize{$\pm0.5$}\\\hline
\end{tabular}
\caption{Few-shot classification accuracies (\%) on CUB-200-2011 and CIFAR-FS.}
\label{tab5}
\end{table}

\subsection{Ablation Studies}
\vspace{-6pt}
%
In this part, we conduct more experiments to analyze the impacts of the designed comparison module, squeeze module, calibration module, multi-head relations, multi-modal class representations.
%
All experiments are conducted on the miniImageNet for the 5-way 1-shot classification task.
\vspace{0pt}

\begin{figure}[t]
\centering
   \includegraphics[width=1.0\linewidth]{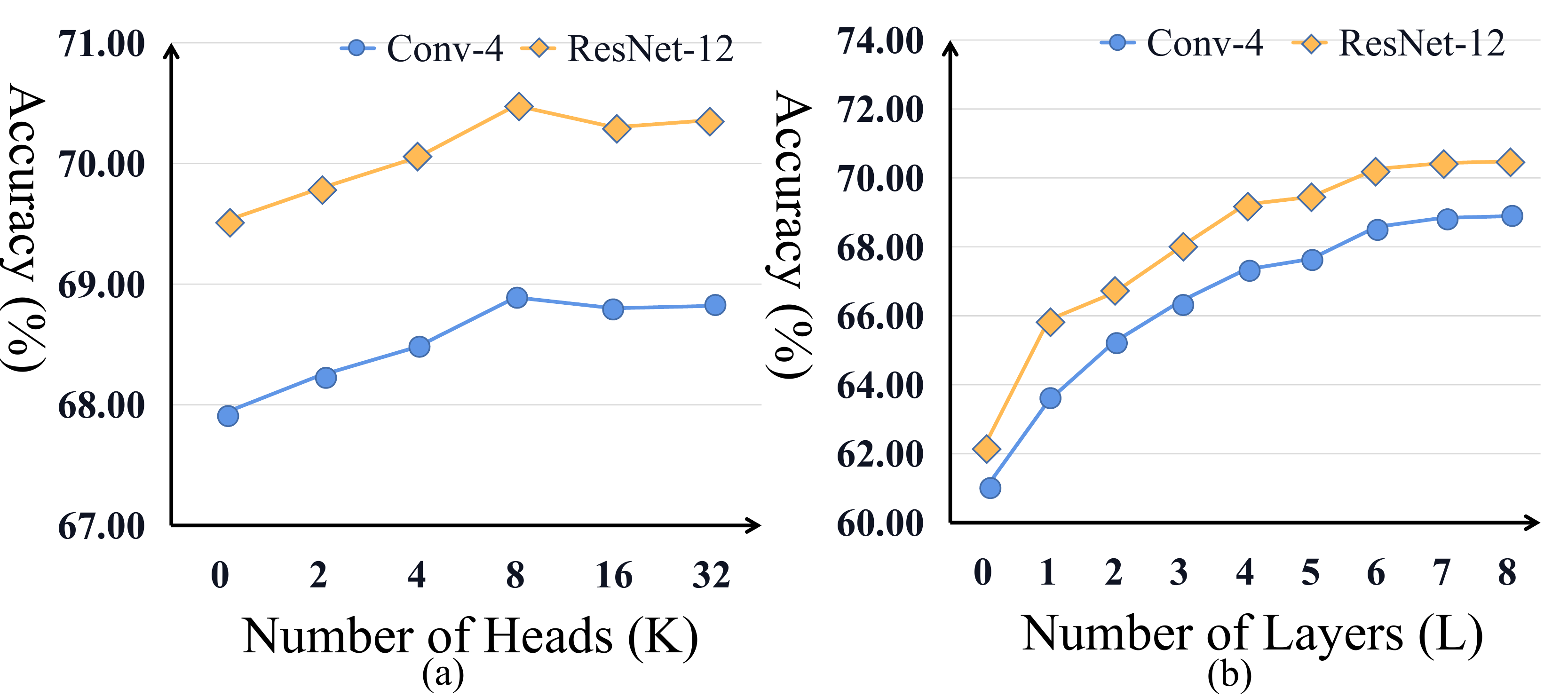}
   \caption{The classification results under different head and layer numbers in 5-way 1-shot on miniImageNet.}
\label{fig:exp2}
\end{figure}

{\bf Impact of the comparison module.}
In the comparison module, we exploit $L$ message passing layers to update the sample representations. In this part, we perform experiments to show the impact of the layer numbers. As shown in Figure~\ref{fig:exp2} (b), with the increase of the layer number, the classification accuracy increases at first and then keeps stable.
Therefore, we set the number of the layers to 6 (i.e., $L=6$) in this paper.
Furthermore, we visualize the similarities of  support and query samples in the first, third and last layers of the comparison module in Figure ~\ref{fig:exp0}.
%
As shown, the proposed ECKPN can characterize the support-query similarities better  with more message passing layers used in the comparison module,
which qualitatively illustrates the effectiveness of the designed comparison module.
%

\vspace{-4pt}
{\bf Impact of the squeeze and calibration modules.} 
In this paper, we design the squeeze and calibration modules to explicitly learn the class-level knowledge to guide the inference of the query samples.
Therefore, it is necessary for us to quantitatively evaluate the effectiveness of these two modules in improving the classification accuracy.
We list the classification results of None-Calibrate and None-Class in Table~\ref{tab6},
where the None-Calibrate denotes the variant of our model without the calibration module, i.e., directly using the class-level knowledge generated in the squeeze module to guide the inference,
and the None-Class denotes the variant of our model without the squeeze and calibration modules, i.e., directly using the pairwise relations in the comparison module for inference.
Compared with the proposed ECKPN,
the classification accuracy of the None-Calibrate decreases by 0.65\% and 0.72\% when using the backbone of Conv-4 and ResNet-12, respectively.
Similarly,
the classification accuracy of the None-class decreases by 1.57\% and 1.36\% when using the backbone of Conv-4 and ResNet-12, respectively.
These results show the effectiveness of the designed squeeze and calibration modules. 

{\bf Impact of the multi-head relations.}
To study the effects of multi-head relations,
we show the classification results of the proposed ECKPN with different head numbers (i.e., $K$ denotes the number of chunks used to separate the visual features.) in Figure~\ref{fig:exp2} (a).
As shown, the head number influences the classification results obviously. To trade-off between the accuracy and complexity, we build the $8$-head relations for message passing in the comparison module. 

{\bf Impact of the multi-modal class knowledge.}
To study the effect of the multi-modal class knowledge, we design two variants of our model, None-Z and None-V.
The former denotes the model without using the semantic knowledge $Z$ (i.e., $V_c^{'}$ in~\eqref{eq:equ18} is equal to $V_c$ ) and the latter denotes the model without using the visual knowledge $V_c$ (i.e., $V_c^{'}$ is equal to $Z$ ).
%
%
As shown in Table~\ref{tab6},
the proposed ECKPN achieves performance gains of 0.3\%-0.5\% and 0.7-0.9\% compared with the None-Z and None-V, which demonstrates the importance of the constructed multi-modal class-level knowledge.
%


\begin{table}
\small
\centering
\begin{tabular}{lcc}
\hline
Method & Conv-4 & ResNet-12\\
\hline\hline
None-Calibrate & 68.24 & 69.76 \\
Non-Class & 67.32 &69.12  \\
Non-Z & 68.53 &69.97  \\
Non-V &68.16 &69.61\\\hline
\textbf{ECKPN} &\textbf{68.89} &\textbf{70.48}\\\hline
\end{tabular}
\caption{The impacts of the squeeze module, the calibration module and the multi-modal class-knowledge in the proposed ECKPN.}
\label{tab6}
\end{table}
\vspace{-6pt}

\section{Conclusion}
\vspace{-6pt}
In this work, we propose a novel Explicit Class Knowledge Propagation Network (ECKPN) for the transductive few-shot classification task.
Our ECKPN stacks three elaborately designed modules of comparison, squeeze and calibration to explicitly explore the class-level knowledge. 
We leverage the generated class-level knowledge representations to guide the inference of the query samples and achieve the state-of-the-art classification performances on four benchmarks, which illustrates the effectiveness of the proposed ECKPN.
In the future, we would like to extend our model for incremental few-shot learning. \\
\textbf{Acknowledgements.} This work was supported by National Key Research and Development Program of China (No. 2018AAA0100604), National Natural Science Foundation of China (No. 61832002, 61720106006, 62072455, 61721004, U1836220, U1705262, 61872424).

{\small
\bibliographystyle{ieee_fullname}
\bibliography{cvpr}
}

\end{document}